\documentclass{article}
\usepackage[final]{corl_2020} 
\usepackage{amsmath}
\usepackage{amssymb}
\usepackage{array}
\usepackage{bbm}
\usepackage{blindtext}
\usepackage{caption}
\usepackage{epsfig}
\usepackage{enumitem}
\usepackage{floatrow}
\usepackage{graphicx}
\usepackage{multirow}
\usepackage{subfig}
\usepackage{wrapfig}

\definecolor{linkpink}{rgb}{0.84705882, 0.05882353, 0.49411765}

\newcommand*\samethanks[1][\value{footnote}]{\footnotemark[#1]}

\graphicspath{ {Figures/} }

\title{\vspace{-1mm}\hspace{-0mm}Robust Policies via Mid-Level Visual Representations: \\
\vspace{2mm}
\Large \hspace{-0mm}An Experimental Study in Manipulation and Navigation}

\author{
  \vspace{1mm}
  Bryan Chen$^1\thanks{Indicates equal contribution.} $\hspace{1mm} 
  Alexander Sax$^{1}\samethanks{}$  \hspace{1mm}
  Gene Lewis$^2$ \hspace{1mm}
  Iro Armeni$^2$ \hspace{1mm} \\

  \vspace{2mm}
  \textbf{
  Silvio Savarese$^2$  \hspace{1mm}
  Amir Zamir$^3$ \hspace{1mm}
  Jitendra Malik$^1$ \hspace{1mm}
  Lerrel Pinto$^{1,4}$ } \\
  
  $^1$ University of California, Berkeley \hspace{1mm}
  $^2$ Stanford University \hspace{1mm}
  $^4$ New York University \\
  
  \hspace{-3.5mm}
  $^3$ Swiss Federal Institute of Technology Lausanne (EPFL) \\
  
}

\begin{document}
\maketitle


\vspace{-1mm}
\begin{abstract}
Vision-based robotics often separates the control loop into one module for perception and a separate module for control. It is possible to train the whole system end-to-end (e.g. with deep RL), but doing it ``from scratch'' comes with a high sample complexity cost and the final result is often brittle, failing unexpectedly if the test environment differs from that of training.

We study the effects of using \textbf{mid-level visual representations} (features learned asynchronously for traditional computer vision objectives), as a \emph{generic} and \emph{easy-to-decode} perceptual state in an end-to-end RL framework. Mid-level representations encode invariances about the world, and we show that they aid generalization, improve sample complexity, and lead to a higher final performance. Compared to other approaches for incorporating invariances, such as domain randomization, asynchronously trained mid-level representations scale better: both to harder problems and to larger domain shifts. In practice, this means that mid-level representations could be used to successfully train policies for tasks where domain randomization and learning-from-scratch failed. We report results on both manipulation and navigation tasks, and for navigation include zero-shot sim-to-real experiments on real robots. \end{abstract}


  

\section{Introduction}
\vspace{-1mm}

Over the past few years, impressive success stories such as~\cite{MnihKSGAWR13, LevineEndToEnd15} have helped deep reinforcement learning (deep RL) make inroads into various fields. Deep RL from pixels, in particular, has drawn attention~\cite{liu2017deep, lazic2018data, silver2016mastering} as a unified method for training agents, but it requires that agents can access virtually unlimited data covering every possible input.

\begin{figure}[H]
    \centering
    \includegraphics[width=0.95\columnwidth]{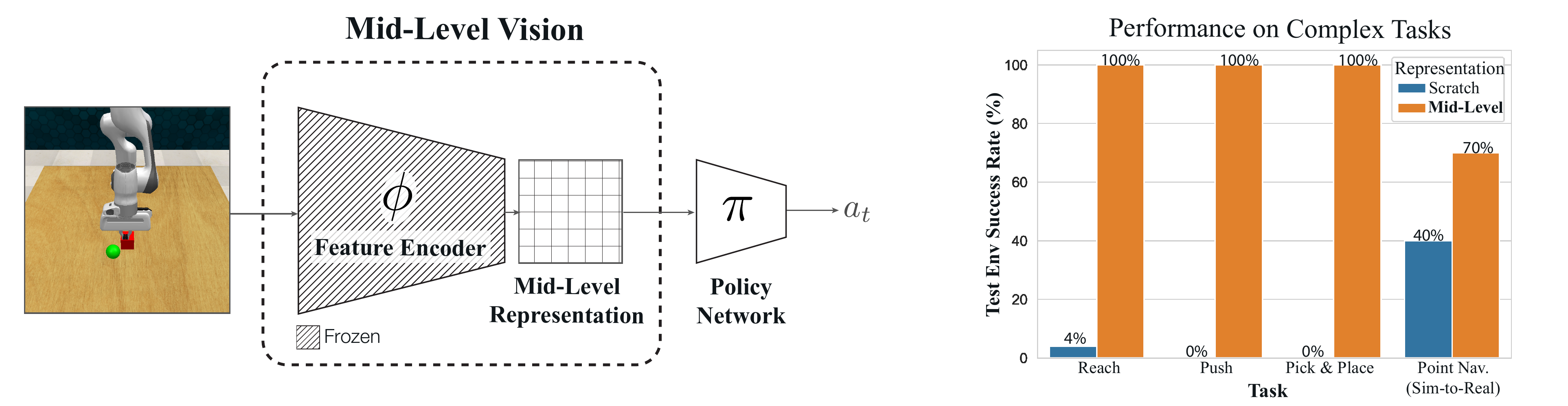}
    \caption{\footnotesize{\textbf{Mid-level visual representations used for RL.} \textbf{Left:} A feature encoder trained for some mid-level objective provides representations to the agent. \textbf{Right:} Agents trained using these mid-level representations were able to generalize, without additional adaptation, to distribution shifts and deployment on physical robots.
    }} 
    \vspace{-3mm}
    \label{fig:pull}
\end{figure}

In scenarios where agents can access large, but still finite amounts of data, the deep-RL-from-pixels approach proves hard to train, however, and the resulting policies are brittle in the real world~\cite{gupta2018robot}. Agents trained this way fail, sometimes spectacularly, under mild visual shifts relative to the training environment. For example, simply placing a water bottle in view of the robot~\cite{LevineEndToEnd15}, or operating in morning light instead of midday sun~\cite{sun2019test,hansen2020self} can be sufficient to completely degrade performance.

This brittleness reflects the fact that the policies have not captured the invariances (and equivariances) necessary to generalize and operate in the real world. The policies do not learn these invariances because there is usually no reason for them to do so. Biased or insufficient training data might permit spurious ``cheating’’ shortcuts or make it easier to memorize features of the training environment rather than learn how to extract those features from the vast space of possible inputs. In any case, agents end up without the right \emph{priors} about the world.

\begin{figure}
    \centering
    \vspace{-0mm}
    \includegraphics[width=1.0\columnwidth]{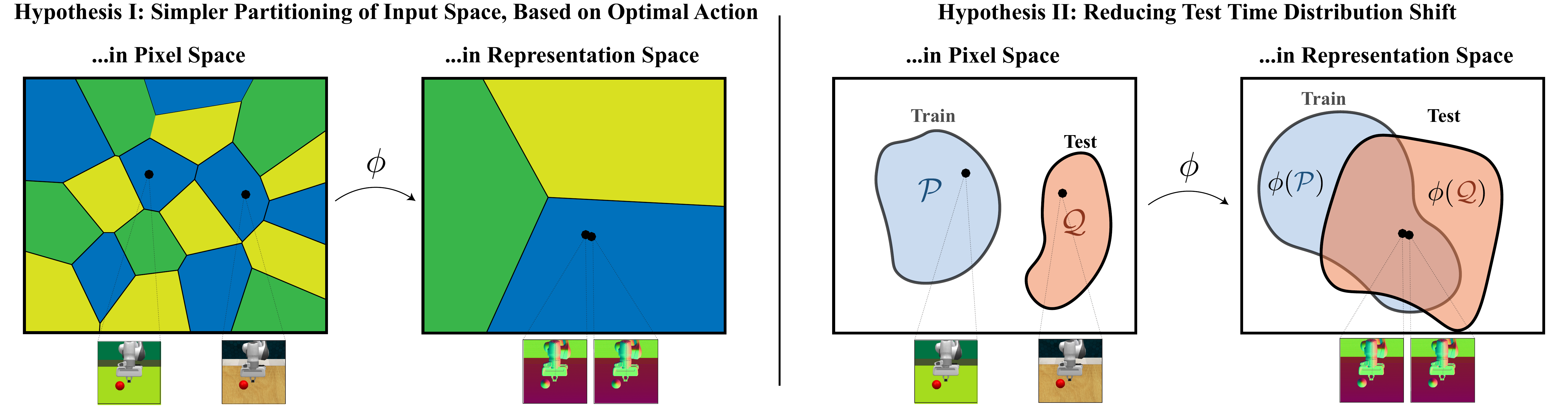}
    \vspace{-1mm}
    \caption{\footnotesize{\textbf{Mid-level representations transform pixel inputs.} There are two major ways that invariances from mid-level representations could be useful for downstream tasks. \textbf{(I) Left:} The invariances simplify the decision boundaries for downstream tasks. In this case, we would expect (i) training on representations to be faster than training on pixels and (ii) to allow us to train agents for more difficult tasks. \textbf{(II) Right:} The invariances in representation space align the train and test distributions. In this case we would expect generalization performance to improve relative to training on pixels. In practice, we see behavior consistent with both hypotheses.
    }} 
    \label{fig:midlevel_vs_downstream}
    \vspace{-1mm}
\end{figure}

Computer vision provides us with tools that parse complex visual scenes and extract usable perceptual representations. In this paper, we study some of those representations used as a form of mid-level vision. That is, instead of training directly on raw pixels, we first extract representations driven by traditional computer vision objectives and use those as the input observations to RL instead (see Fig.~\ref{fig:midlevel_vs_downstream}-left). The networks which extract the mid-level visual representations are \emph{asynchronously trained}, meaning that they can be trained independently and on a schedule different from the RL training. This approach has shown promise~\cite{sax2019midlevel,vladlen2019,yenchen2020plagiarism}, especially in navigation contexts where agents using mid-level vision are able to generalize to unseen (simulated) buildings~\cite{sax2019midlevel}. 

The main contribution of our study is experimental. We show that this approach scales to harder tasks than previously shown, even when control is nontrivial (e.g. manipulation with continuous control), or there are drastic domain shifts such as training in simulation and then deploying on physical robots with no additional training. Specifically, we test the following hypotheses:
\vspace{-1mm}
\begin{enumerate}[leftmargin=7mm,label=\alph*]
    \vspace{-1mm}
    \item Do mid-level visual representations provide a useful way to incorporate invariances for ``hard" tasks, when compared to training from scratch? (Answer: \textbf{Yes})
    \vspace{-1mm}
    \item Do the representations simplify the learning problem (opening up the possibility to successfully train on harder problems that fail otherwise)? (Answer: \textbf{Yes})
    \vspace{-1mm}
    \item Do the representations improve robustness to distribution shifts? (Answer: \textbf{Yes}: both sim-to-real and within simulation)
    \vspace{-1mm}
    \item How does the mid-level approach compare to other approaches for incorporating invariances? (Answer: \textbf{It scales significantly better})
\end{enumerate}
\vspace{-1mm}

In summary, we present a large-scale study evaluating the effect of using invariances learned from computer vision objectives plugged into active RL frameworks. We find that the mid-level approach actually performs even better in the harder contexts, relative to alternatives, and provides an avenue for solving harder problems. We analyze this behavior in terms of training performance, generalization performance, and sample complexity. The mid-level approach was able to achieve a 100\% success rate in certain test environments when alternatives approaches like learning-from-scratch, using  \emph{ground-truth low-dimensional state}, and domain randomization do not learn at all (0\% train and 0\% test, even after multiple hyperparameter sweeps).The mid-level approach trains faster than the alternatives, almost as fast as using ground truth low-dimensional state (when state succeeds). Further, we compare mid-level representations to other methods of learning invariances (domain randomization), and show that as the tasks become more difficult, domain randomization makes learning the task harder ($100\%$ train success $\rightarrow70\%$  with domain randomization, $4\%\rightarrow20\%$ test) while mid-level simplifies it (near-perfect performance on the both train and test domains).


\section{Related Works}
\label{sec:related}
\vspace{-1mm}

Our study is connected to a range of relevant fields and we review the most important ones, within space constraints.

\textbf{Computer Vision.} Computer vision approaches are typically designed to solve stand-alone vision objectives such as depth estimation~\cite{EigenPF14}, object classification~\cite{alexnet}, detection~\cite{Girshick15}, segmentation~\cite{Silberman2012}, pose estimation~\cite{iss, Cao2018openpose}. While approaches may use various levels of supervision~\cite{alexnet, NorooziF16, doersch2015unsupervised, bengio2013representation,pinto2016curious}, the common characteristic across conventional computer vision methods is that they are trained and evaluated on static datasets collected for that stand-alone objective. In contrast, the perception of active agents is fundamentally in service of some downstream goal, and agents are evaluated on that goal in an \emph{online} manner so that a single decision early on in an episode impacts the observations that will follow. In this paper we study how conventional computer vision objectives impact these downstream tasks, especially when the downstream tasks have strong temporal dependencies and domains different than those used for training the conventional vision approaches.

\textbf{Representation/Feature Learning.} The goal of representation learning is to encode observations in a way that provides benefits over using raw sensor data. One of the most popular approaches is based on Minimum Description Length (MDL) which suggests good representations are those which most compactly describe the data. This includes variational autoencoders~\cite{kingma2013auto} and alternatives~\cite{Hinton504, Vincent:2008:ECR:1390156.1390294, Matthey2017betaVAELB} and has been applied to robotics (e.g. ~\cite{finn2016deep}). However, MDL representations tend to be exceedingly sensitive to tiny domain shifts~\cite{sax2019midlevel}, an obervation that we also make. Likelihood-based approaches often try to predict the probability that two patches come from the same image--e.g. Constrastive Predictive Coding (CPC)~\cite{vandenOord2018predictive} and variants~\cite{cpcpp2019, chen2020simple, he2019momentum, grill2020bootstrap}. Dynamics-based approaches model the environment (e.g. by predicting the next state ~\cite{Jordan1992ForwardMS} or related objectives~\cite{Dosovitskiy16predicting, curiosity, zhu2007, Raffin2018srltoolbox, AgrawalNAML16, yan2020learning}). Dynamics models have the possible advantage that they could be reused for planning. Dynamics are not necessarily visual and are sometimes specialized to the morphology or action space of the agent.

\textbf{Mid-Level Vision in Other Contexts.}
Recent works have shown that mid-level visual representations can offer advantages in terms of generalization and sample complexity for downstream active tasks. Many works show one or a couple objectives for single tasks--e.g. a specific semantic representation for semantic driving~(\citep{modularityAbstraction, mousavian18, Yang2018ScenePriors}) or dense object descriptors for manipulation~(\citep{denseobjectFlorence2018}). Outside of RL,~\cite{yenchen2020plagiarism} uses mid-level vision in conjunction with hard-coded grasping policies.

Recently, \cite{sax2019midlevel, vladlen2019} presented comprehensive studies in simulated navigation contexts using multiple objectives and multiple tasks. This many tasks/many objectives setup allows them to conclude that which objectives perform well depends on the task, which we also find. \cite{sax2019midlevel} then computationally derives a generic subset of mid-level representations that should perform as  well as the best one in the whole set, but does not show when that best-possible feature is useful. Unlike previous studies, we demonstrate mid-level vision's ability to scale to harder tasks, to handle sim-to-real transfer, and we compare it to other approaches for incorporating invariances (e.g. domain randomization). By examining more complex domains, we find that we are able to train agents using mid-level features even in cases where other approaches fail.

\textbf{Domain Randomization.}
One way of building agents with useful invariances is to make them learn it directly from data. \cite{tobin2017domain,sadeghi2016cad2rl,pinto2017asymmetric,wu2019learning,andrychowicz2020learning} suggests using simulators to augment training with all the variation likely to be present at testing. However, training then takes longer and, in practice, varying more than one or two factors complicates the learning problem to the point that learning completely fails. Most importantly, domain randomization involves making the unrealistic assumption that we can enumerate all the invariances that are needed. We cannot, and for those which we can define, the corresponding variation is often difficult to build into a simulator (as evidenced by the numerous open problems in computer graphics). Our study shows that incorporating invariances via mid-level representations has multiple critical advantages over doing so via domain randomization. In particular, the mid-level representations can be trained asynchronously on large static datasets, making RL training simpler and more scalable. In contrast, domain randomization (unnecessarily) delays learning the invariances to be done in conjunction with RL training. 




\section{Methodology}
\label{sec:method}

Our goal is to study the utility of mid-level visual representations for performing downstream motor tasks. Concretely, motor tasks require mapping observation histories to actions. Agents are trained in a scenario $\mathcal{P}$ but evaluated on a test scenario $\mathcal{Q}$. The goal is to maximize agent's performance in a test setting ($\mathcal{Q}$) that it is related, \emph{but not identical}, to the training distribution. For example, $\mathcal{Q}$ might be a physical robot in scenes contain unseen objects, while training on $\mathcal{P}$ took place in a simulator. 

We assume access to a set of `mid-level' functions, $\Phi = \{\phi_1, \ldots, \phi_m\}$, that transform raw sensory data into potentially useful mid-level representations. The goal is to determine whether agents using $\Phi$ could perform better in the test setting $\mathcal{Q}$ than if they never had access to $\Phi$. In this paper, we show that when $\Phi$ is a set of mid-level functions, agents access to $\Phi$ improves generalization and generalize better final perfomance.

\subsection{Using Mid-Level Representations in Active Contexts}

Why might a mid-level feature (e.g. image $\rightarrow$ surface normals) aid in downstream tasks, compared to end-to-end learning? This is an important question as preprocessing observations with $\phi$ might potentially discard information (e.g. color information in surface normals). Good representations preserve important information while discarding spurious details, providing ``invariances" that make the train and test set more similar ($\phi(\mathcal{P}) \approx \phi(\mathcal{Q})$). When the train and test set become similar, improving performance on the train set generally improves test-time performance, too.

\emph{Really} good representations also simplify training by using $\phi(\mathcal{P})$ instead of $\mathcal{P}$. By throwing away unimportant information and providing easily decodable outputs (e.g. linearly separable), great representations can reduce sample complexity and boost performance even on the training set, relative to learning \emph{tabula rasa}. These ideas are illustrated in Figure~\ref{fig:midlevel_vs_downstream}.

We use representations derived from neural networks trained, offline, for computer vision objectives. Figure~\ref{fig:pull} shows how we use these networks in our active framework. While training agents from mid-level representations we freeze the mid-level network ($\phi$) and use it to transform each observed image $o_t$ into a summary statistic $\phi(o_t)$ that is then provided to the agent instead of pixels. Only the policy network is updated during agent training. 
This has the advantage that we can reuse the same mid-level features for new tasks without degrading the performance of already-learned policies. Freezing the representations is a na\"ive approach and the features could be updated (e.g. via~\cite{rebuffi2017learning, zhang2019sidetuning}), but even with the na\"ive approach the representations yield notable benefits.

\begin{figure}
    \centering
    \vspace{-6mm}
    \includegraphics[width=\columnwidth]{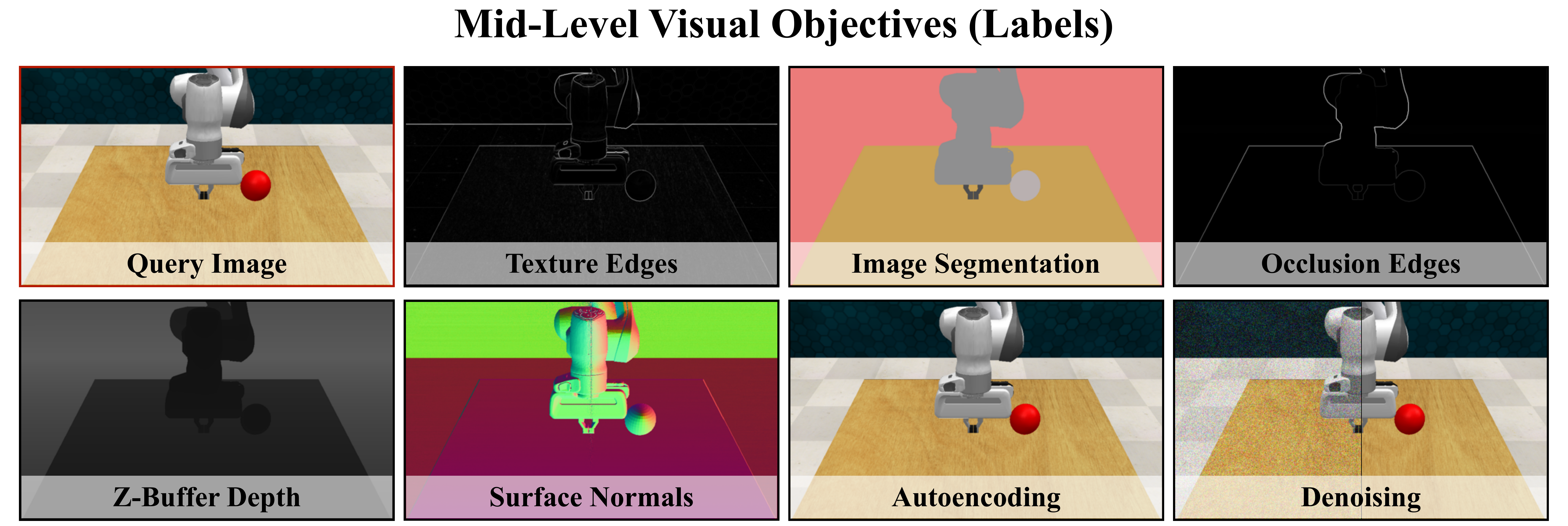}
    \caption{\footnotesize{\textbf{Sample labels for mid-level visual objectives in the RLBench environment.} The objectives cover various modes of computer vision tasks including 2D, 3D, and semantic tasks.}} 
    \vspace{-2mm}
    \label{fig:representation_viz}
\end{figure}

\textbf{Studied Mid-Level Representations}
In this paper, we study representations from neural networks that were trained, offline, each for a specific vision objective from~\cite{taskonomy2018}. 7 of them are visualized in (Figure~\ref{fig:representation_viz}) and they cover various common modes of computer vision objectives: from texture-based (e.g. denoising), to 3D pixel-level (e.g. depth estimation), to semantic (e.g. image segmentation). As the networks were trained on data from indoor scenes significantly different than our manipulation setting, we fine-tune the networks, offline, with images from our simulator. We study the effects of domain shift and feature robustness in Section~\ref{sec:analysis}. All networks were trained with identical hyperparameters and using a ResNet-50~\cite{resnet} encoder. For a full list of vision objectives and samples of the networks evaluated in our environments, see the supplementary.

\subsection{Metrics}

\textbf{Final performance:} We report train and test performance at the end of training. That not all agents converge to the same training reward may be due to the limitations of current learning approaches, data and compute; an agent might still fail to learn \emph{anything} even when it is given every available resource. We find this occurs regularly for complex tasks, even when agents can access the true low-dimensional environment state (see Section~\ref{sec:generalization}). Specific mid-level features might also be ill-suited to particular tasks; discarding necessary information and further limiting final performance. 

\textbf{Generalization:} We break down final performance in terms of generalization from train to various test sets. We report both test-set performance with no shift (evaluated in the same simulator as training), under mild domain shift (with various colors swapped out at test-time), and under more drastic domain shift (unseen objects or under zero-shot sim-to-real transfer).

\textbf{Sample efficiency:} We examine whether agents equipped with mid-level vision can learn faster than a comparable agent learning \emph{tabula rasa} (i.e. with raw vision, but no priors about the world). We report this in terms of the number of interactions with the environment. 

\subsection{Manipulation with Continuous Control}

To study mid-level vision on harder tasks, we test the mid-level representations against baselines in three different manipulation tasks requiring continuous control from vision. This section describes the setup at a high level, and complete details for each subsection can be found in the supplementary.

\textbf{Tasks:} We use three common manipulation tasks: \emph{Reach}, \emph{Push}, and \emph{Pick + Place} with sparse rewards (0 if within $\epsilon$ of the target, -1 otherwise). All tasks terminate episodes after 50 timesteps. 
Brief descriptions of each task are below, and full details for each of the the tasks are in the supplementary.
\begin{description}[leftmargin=3mm]
\small
\vspace{-1mm}
\item \textbf{Reach:} The agent must move the end effector to a random target position marked by a sphere. 
\vspace{-1mm}
\item \textbf{Push:} The agent must push a cube to a random target position marked by a sphere.
\vspace{-1mm}
\item \textbf{Pick and Place}: The agent must pick up a cube and move it to a target position marked by a sphere. Both object and target locations are randomized.
\end{description}

\textbf{Observation Space:} For each task agents receive visual input from a single camera. Agents trained from pixels receive $64{\times}64{\times}3$ RGB images while mid-level approaches receive $16{\times}16{\times}8$ latent features. Except for the state baselines, no other information (including proprioception) is supplied.

\textbf{Action Space:} For all tasks, actions are XYZ+gripper values in [-1, 1] specifying end-effector deltas for a position controller. Gripper open/close is calculated via thresholding.

\begin{figure}
    \RawFloats
    \centering
	\begin{minipage}[]{0.38\textwidth}
        \vspace{-0.2mm}
        \includegraphics[width=\columnwidth]{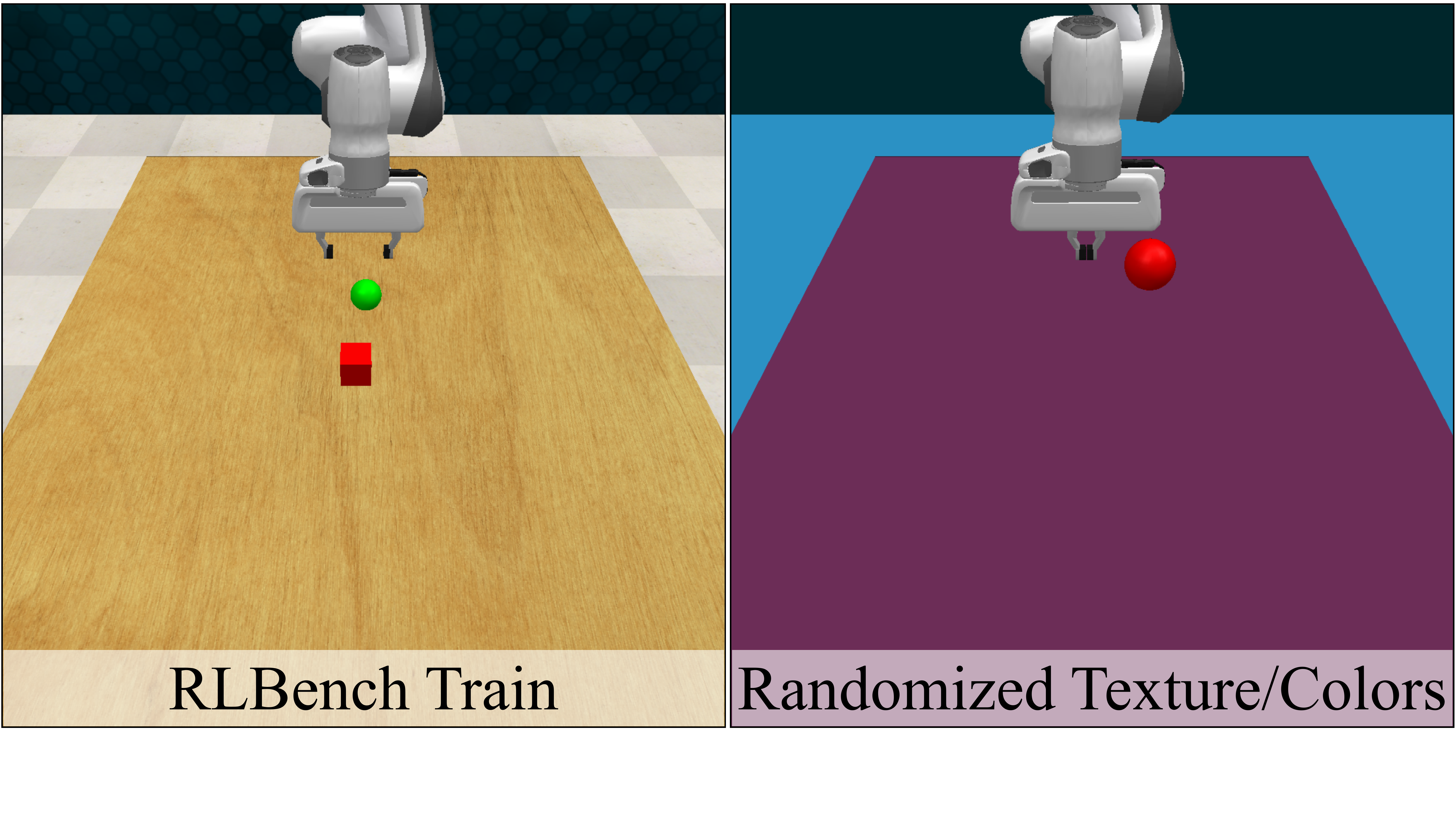}
        \vspace{-5mm}
        \captionof{figure}{\footnotesize{\textbf{Train vs. test observations in RLBench.} \textbf{Left:} The default environment for the Pick + Place task. The goal is colored green, object in red. \textbf{Right:} Sample observation from the test environment showing held-out randomized textures.
            }} 
        \vspace{-4mm}
        \label{fig:rlbench}
    \end{minipage}
    \hfill
	\begin{minipage}[]{0.58\textwidth}
        \centering
        \vspace{-5.5mm}
        \includegraphics[width=0.98\textwidth]{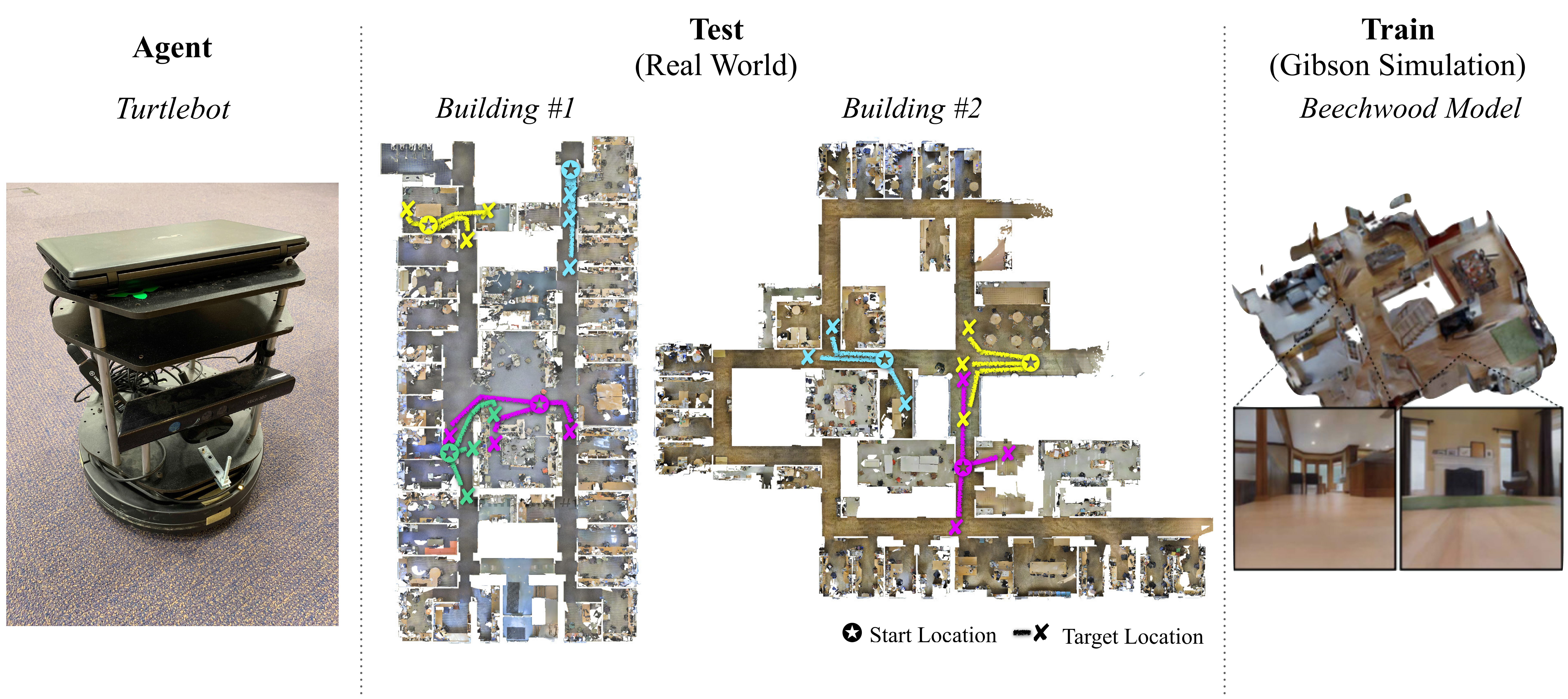}
        \vspace{-1.5mm}
        \caption{\footnotesize{\textbf{Zero-Shot Visual Sim-to-Real.} \textbf{Right:} We train policies in a single building in simulation. \textbf{Middle:} We then directly test them in novel real world buildings, where agents have no prior knowledge of the building and no adaption period. \textbf{Left:} The TurtleBot uses only an RGB camera for vision and an IMU for localization, No depth/LiDAR sensors are used.}}
        \label{fig:qual}
        \vspace{-4.0mm}
    \end{minipage}
\end{figure}

\textbf{Environment:} We adopt the RLBench environment~\citep{james_rlbench_2019} which is built on PyRep~\cite{pyrep} and CoppeliaSim~\cite{coppeliaSim}. RLBench is suited for vision-based manipulation and offers more visually realistic observations compared to popular environments such as OpenAI Gym \cite{fetch}. We use the Franka Emika Panda arm in our experiments.   

\textbf{Train/Test Split}
Agents are trained in the default RLBench tabletop environment shown in Figure~\ref{fig:rlbench}. All policies are tested on a test set of held-out flat color textures that replace the table, floor, and/or background (Fig~\ref{fig:rlbench}, right). We also show experiments with domain randomization during training (no texture overlap with the test set) or with held-out objects during testing.

\textbf{Learning Algorithm.} 
We train all agents using TD3~\cite{TD3} and HER~\cite{HER} using hyperparameters optimized for the \emph{tabula rasa} baseline. Mid-level agents used the learning rate optimized for scratch in \cite{sax2019midlevel}, but we ran additional hyperparameters sweeps for agents trained from scratch on the Reach and Pick+Place tasks. Because all hyperparameter searches were done using the from-scratch approach and the settings then applied everywhere, this setup should be biased \emph{against} mid-level vision.

We followed the guidelines for training laid out in~\cite{HER}. For some tasks we could not get the from-scratch approach to train using sparse rewards and we had to add additional reward shaping. The shaped dense rewards often help the mid-level approach too (see supplementary). While in the main paper we present the best approach for the pixel-based methods, we only show the sparse-trained policies for mid-level based agents since sparse rewards are usually easier to define in practice.

\subsection{Generalization to the Real World}

In cases where collecting real-world data makes training policies prohibitively expensive, the plentiful and cheap data from simulation provides a path to train policies that can then be deployed in the real world. However, the domain shift between simulators and the real world means that policies trained this way usually fail to generalize. We study this sim-to-real capability for agents trained with mid-level representations via RL.

As we are primarily focused on perception, we focus on navigation contexts where the visual gap is responsible for the primary domain shift. Manipulation is usually non-quasistatic and, in practice, simulators will trade off accuracy (simulating all contact forces) for simulation throughput, resulting in a large sim-to-real gap for environment dynamics. Simulators can more accurately model the simpler dynamics of navigation environments and physical robots have good low-level locomotion controllers that can handle any discrepancies. Even so, navigation-based sim-to-real is highly non-trivial from a vision perspective. Any number of discrepancies between simulated images and real images could cause the agent to fail to generalize; potential discrepancies include lighting variations, stitching artifacts and semantic distribution complexity.

We test sim-to-real generalization for the \emph{point navigation} task from the CVPR19 Habitat Challenge. The task requires an agent to navigate to a target position (specified by coordinates) using visual observations and the agent's onboard odometry. Actions are discrete \texttt{\{forward, pivot right, pivot left\}} and episodes cap at 400 timesteps. We use policies from~\cite{sax2019midlevel} trained with the same architecture as our manipulation tasks but using PPO~\cite{PPO} in a \emph{single} building in the Gibson environment~\cite{gibson} and are tested in \emph{different} (unseen) real-world buildings. We evaluate the mid-level vision based policies trained in~\cite{sax2019midlevel} on a Turtlebot with Kobuki base and a Microsoft Kinect camera. 
Full details are provided in the supplementary.

\subsection{State Representation Baselines}
\vspace{-1mm}
To address confounding factors, we compare against several controls. We describe the most important ones here and defer remaining descriptions to the supplementary.
\vspace{-1mm}
\begin{description}[leftmargin=3mm]
    \vspace{-1mm}
    \small
    \item \textbf{Tabula Rasa Learning (aka \emph{from scratch}):} This is the most common approach for end-to-end learning. The agent receives the raw RGB image as input and uses a randomly initialized AtariNet~\cite{MnihKSGAWR13} architecture that is updated during training, along with the policy architecture.
    \vspace{-1mm}
    \item\textbf{Blind:} The blind agent is the same as \emph{scratch} but instead of an RGB image, receives a constant zero tensor. This shows how much can be learned by just exploiting the biases of the task.
    \vspace{-1mm}
    \item\textbf{State:} In this setting the agent has access to the complete environment state: joint positions, the goal centroid and, if applicable, the object center. Since objects are always the same, this should be sufficient.
\end{description}


\section{Results}
\label{sec:results}
Given that no two setups outside of simulation will be exactly the same, building in invariances into visuomotor policies could be helpful for bridging the gap. The rest of this section covers experiments that dissect whether such invariances are necessary, finding that incorporating them offers notable advantages in terms of final performance, generalization, and sample complexity. We then compare two main methods of building in invariances: either co-learning them during training (via domain randomization) or asyncronously learning them in different stages (via mid-level representations). As the agent needs to learn more invariances, we show that the co-learning becomes complicated and learning can collapse. We find that the mid-level approach scales and performs better.

\subsection{Final Performance}

\begin{figure}[H]
    \centering
    \vspace{-1mm}
    \includegraphics[width=0.7\columnwidth]{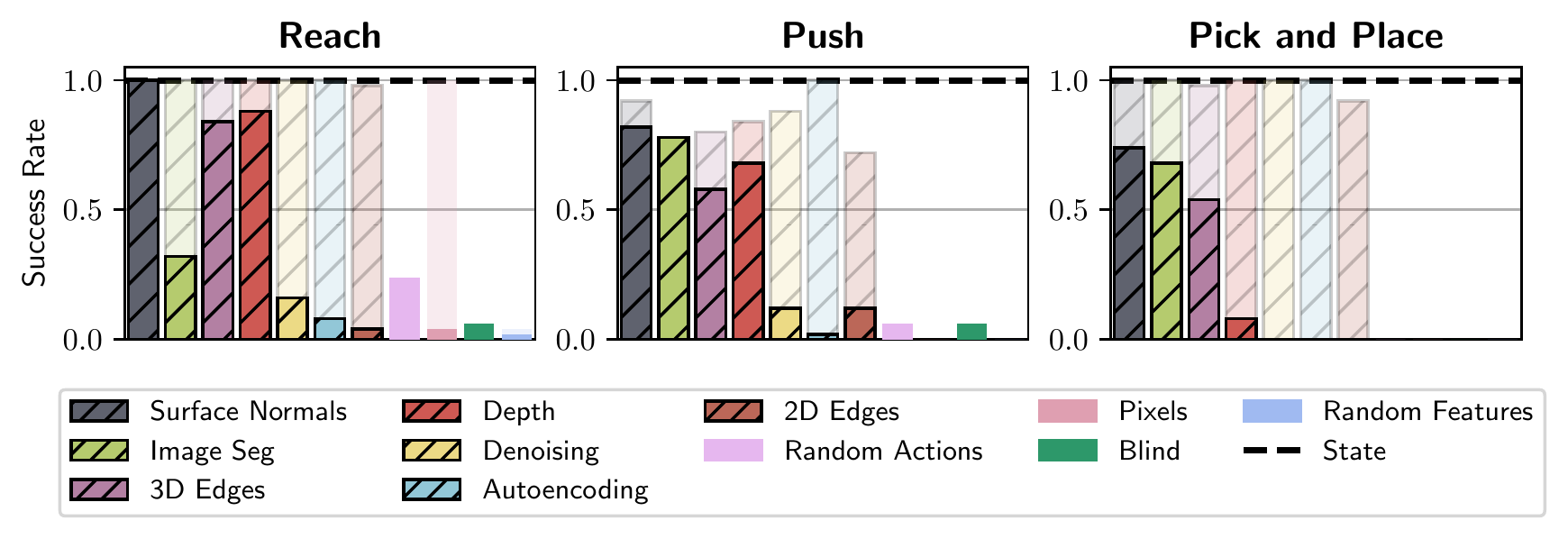}
    \vspace{-2mm}
    \caption{\footnotesize{\textbf{Final performance on manipulation tasks.} Translucent bars indicate training performance; opaque bars show performance on the test set. Hatched bars indicate agents are trained with mid-level representations. 
    }} 
    \label{fig:final_performance}
\end{figure}

We find that agents trained using mid-level representations achieve significantly higher success rates than agents learning from pixels, especially in harder tasks such as \emph{Pick + Place}. The mid-level agents performed much better than scratch in the test environment,  as shown in Fig.~\ref{fig:final_performance}. Consistent with Hypothesis II (\emph{mid-level representations simplify training}) they performed better during training, too, especially for harder tasks.\footnote{The mid-level policies in Fig.~\ref{fig:final_performance} used vision networks fine-tuned from~\cite{taskonomy2018}. When they were retrained for longer (and from scratch), performance for the mid-level agents improved further and they roughly matched the performance of the \emph{state} oracle (Section~\ref{sec:generalization}). Results shown in Table~\ref{fig:sim_to_sim}.}

\vspace{-4mm}
We found the above results despite the fact that the hyperparameters were optimized for the scratch baseline. In general, we found that the mid-level agents were less sensitive to choices of hyperparameters, and that the same or similar hyperparameters worked across multiple architectures, downstream learning algorithms, and simulators. They also did not require reward shaping.


\subsection{Generalization}\label{sec:generalization}

\begin{figure}[H]
        \RawFloats
		\begin{minipage}[]{\textwidth}
			\hspace{-0.0mm}\begin{minipage}[]{0.4\textwidth}
				\centering
				\vspace{-0.0mm} \hspace{-0.0mm}\includegraphics[width=1.0\columnwidth]{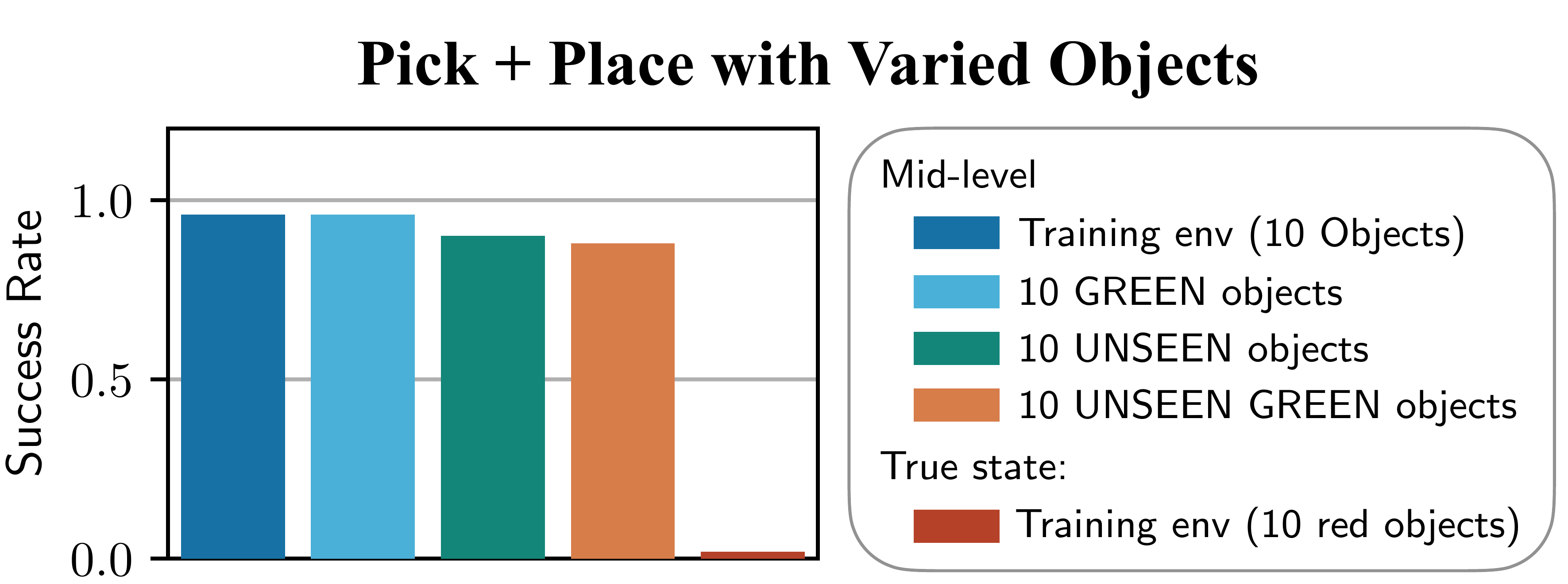}
				\vspace{-2.3mm}
				\captionof{figure}{ \footnotesize{\textbf{Generalization to unseen objects.} In all tested environments, the mid-level policy trained to pick+place varied objects outperformed an agent using true state.}
				}
				\label{fig:procedural_exps}
				\vspace{-1mm}
			\end{minipage}
			\hfill
            \hspace{2.5mm}
			\begin{minipage}[]{0.575\textwidth}
				\centering
				\scriptsize
				\setlength{\tabcolsep}{4pt}
				\begin{tabular}{|l|c|ll|ll|}
                \hline
                \multicolumn{2}{|c|}{\underline{Invariance Method}} & \multicolumn{4}{c|}{\underline{{Success Rate}}} \\
                \multicolumn{1}{|c|}{\textbf{Mid-Level}} & \multicolumn{1}{c|}{\multirow{2}{*}{\textbf{Domain Rand.}}} & \multicolumn{2}{c|}{\textbf{Reach}} & \multicolumn{2}{c|}{\textbf{Pick + Place}}  \\
                \multicolumn{1}{|c|}{\textbf{Representation}} && \multicolumn{1}{c}{Train} & \multicolumn{1}{c|}{Test} & \multicolumn{1}{c}{Train} & \multicolumn{1}{c|}{Test}  \\ 
                \hline
                \hline
                None (\textit{Scratch})       &  \multirow{3}{*}{No}    & 100\%   & 4\% & 0\%  & 0\% \\
                Image Segmentation & & 100\%  & 88\% & 100\%  & 92\% \\
                \textbf{Surface Normals} &    & \textbf{100\%}  & \textbf{100\%}  & \textbf{100\%}  & \textbf{100\%} \\
                \hline
                None (\textit{Scratch})        &  \multirow{3}{*}{Yes} & 70\%   & 20\%  & 0\%  & 0\% \\
                Image Segmentation & & 100\%  & 100\% & 98\%  & 98\% \\
                \textbf{Surface Normals} &    & \textbf{100\%}  & \textbf{100\%}  & \textbf{100\%}  & \textbf{100\%} \\
                \hline
            \end{tabular}
            \captionof{table}{
                \footnotesize{\textbf{Policies trained via different invariance-learning approaches.} The mid-level approach scales better to harder tasks, compared to \emph{tabula rasa} or domain randomization.}
            }
            \label{fig:sim_to_sim}
		\end{minipage}
	\vspace{-4mm}
	\end{minipage}
	\end{figure}

\textbf{Comparison to generic state for unseen objects.} 
In order to test whether mid-level representations could provide an easily decodable representation that enables both learning and generalization to unseen objects, we train agents for Pick+Place with 10 red, procedurally generated objects of different shapes. In contrast to the standard environment which only used a red cube, encoding the salient parts of the environment is now more complicated; we represent the object shape by its mesh vertex positions centered by the object centroid. In the more complex training environment, the state-based agent gets a 2\% success rate during training (0\% on unseen test), shown in Fig.~\ref{fig:procedural_exps}. In contrast, using mid-level representations (normals), the agent has a 96\% success rate on the training objects (90\% on the unseen objects, 96\% training objects colored green, and 88\% on unseen green objects). 



\textbf{Learning invariances with mid-level representations vs. domain randomization}. Table~\ref{fig:sim_to_sim} compares the performance of agents trained with different methods of incorporating invariances. Agents using the asyncronous (mid-level) approach perform better across train and all test environments. In particular, when tested on colors not seen in the training, mid-level vision has a success rate of 100\% versus 20\% when using pixels with domain randomization. The domain randomization approach trained from scratch also showed signs of learning collapse (100\% $\rightarrow$ 70\% success rate) as the randomization made the learning problem more difficult, a problem also found in~\cite{james2019sim, akkaya2019solving}.

\begin{wrapfigure}{r}{0.5\textwidth}
	\begin{minipage}[]{1.0\textwidth}
        \centering
        \vspace{-5.0mm}
        \includegraphics[width=1.0\columnwidth]{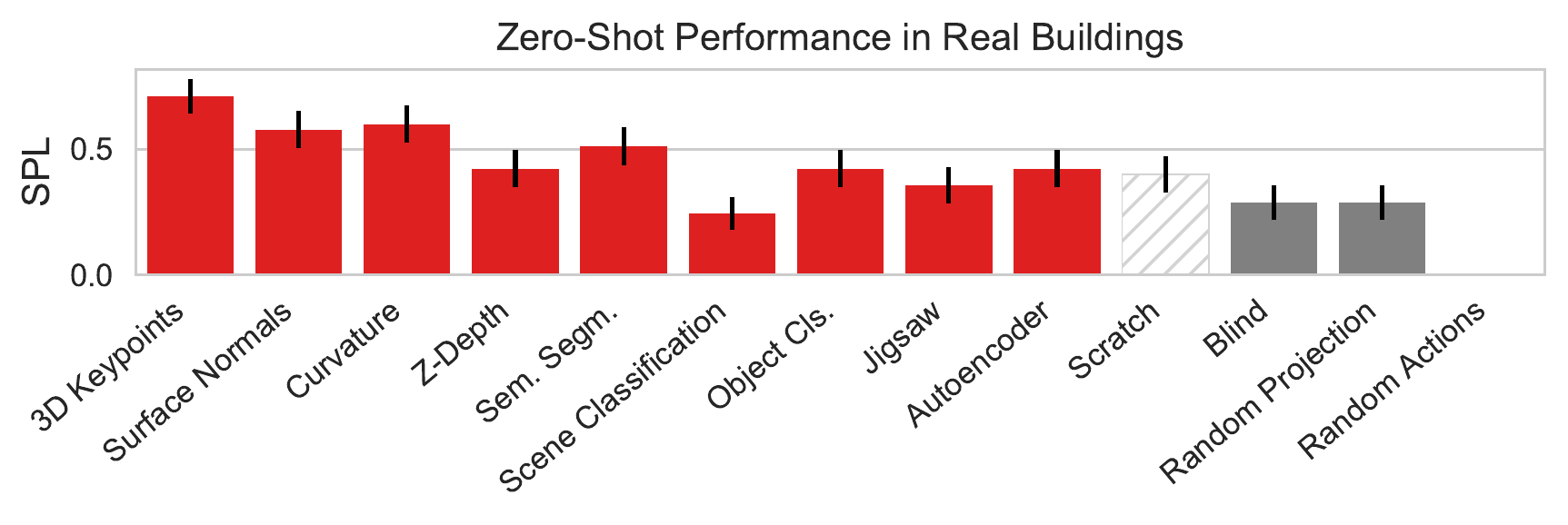}
        \vspace{-6mm}
        \captionof{figure}{\footnotesize{\textbf{Performance in simulation vs. zero-shot transfer to the real world.} Agents using mid-level representations (red) significantly outperform agents trained from scratch. Agents are ordered on the x-axis by descending performance in simulation. Those trained from scratch do not significantly outperform blind agents in the real environment (standard error shown in chart). Performance for top-performing features (e.g. curvature) was about the same in simulation as the on a physical robot~\cite{sax2019midlevel}.
        }} 
        \vspace{-2mm}
        \label{fig:sim_to_real_performance}
    \end{minipage}
\end{wrapfigure}

\textbf{Sim-to-real transfer.} Agents using mid-level vision generalize to new axes of variation not present during training and across large gaps of the simulator vs. physical world. After training in a single simulated building, we test in 24 scenarios in two unseen buildings in the real world. Scenarios vary significantly in length, complexity, and visual characteristics (mean length 5.24m; variance 3.65$m^2$). In 594 evaluation runs and over 13 hours of execution time, we found that agents trained from scratch achieved an SPL~\cite{anderson2018evaluation} of 0.319 and a completion rate of 0.4 in the test environment, which was not significantly different than blind agents, as shown in Figure~\ref{fig:sim_to_real_performance}. 
Agents using mid-level features achieved a significantly higher SPL of 0.608 and a completion rate of 0.7. The use of the best features therefore provides a 90.6\% increase in SPL and 75.3\% increase in completion rate over scratch. Because we did no fine-tuning here, we could evaluate a slightly larger set of features here than for the simulation experiments. A full description of the experiment is available in the supplementary, and we provide videos from the agents' onboard cameras during the sim-to-real test episodes on our \href{http://midlevel.berkeley.edu}{website}.

\subsection{Sample Efficiency}




Training agents with mid-level representations dramatically increases sample efficiency, allowing for policies to be trained on difficult tasks using sparse rewards. Across all tasks, agents using mid-level representations converge within 2x the number of steps required to train from state (e.g. 450k steps vs 250k in Pick + Place). This is several times faster than learning from scratch (when it is even possible for scratch to learn anything---even with reward engineering, the from-scratch approach never succeeds on the test set). We provide train/test curves in the supplementary.

\subsection{
Analysis of 
Features for Downstream Tasks}\label{sec:analysis}

\textbf{Relationship between mid-level objectives and downstream tasks.}

\begin{wrapfigure}{r}{0.5\textwidth}
    \RawFloats
    \vspace{-5mm}
    \begin{minipage}[]{1.0\textwidth}
        \centering
        \vspace{0mm}
        \includegraphics[width=1.0\columnwidth,]{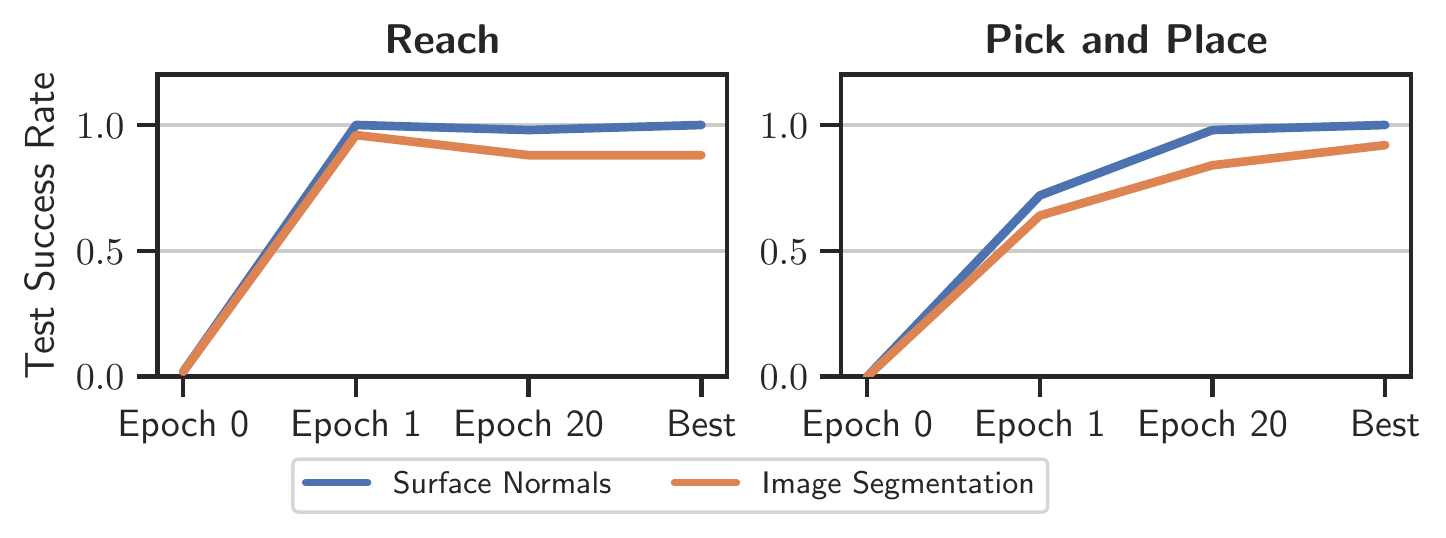}\\
        \captionof{figure}{\footnotesize{\textbf{Performance on mid-level vs. downstream tasks.}  If a feature initially performed well on the downstream task, a better version further improved downstream performance. 
        }} 
        \vspace{-0mm}
        \label{fig:midlevel_vs_downstream_analysis}
    \end{minipage}
    \vspace{-1mm}
\end{wrapfigure}

Given that mid-level objectives are typically defined irrespective of any downstream task, we ask whether representations that perform better on their objective also perform better on downstream tasks. Figure~\ref{fig:midlevel_vs_downstream_analysis} shows the downstream performance of agents trained using surface normal and image segmentations features when those features are from various checkpoints during training. We found that generally, when the feature is useful for the task, the two performances are correlated (both features on pick+place). 



\textbf{Feature rank stability in different environments.} 

\vspace{-0mm}

We found that within each task, feature order was notably stable, even across large sim-to-real domain gaps: the Spearman's rank correlation between the feature rankings found by testing in the real world vs. testing in simulation (Gibson) was $\rho{=}0.77$. \cite{sax2019midlevel} compares across simulators, but not in a zero-shot manner, finding that feature rank correlation was $\rho{=}0.88$ between Habitat~\cite{habitat19iccv} and Gibson when mid-level agents were trained in the respective environments. This inter-environment correlation is not simply because some features are more useful than others (i.e. the same features are always useful): feature ranking in \cite{sax2019midlevel} was uncorrelated across tasks.


\hfill \break
\hfill \break
\section{Conclusion}
\label{sec:conclusion}
We presented a comprehensive experimental study on using mid-level visual representations with RL to train agents to complete complex tasks over varying levels of distribution shifts. 

\noindent\textbf{High-level takeaway:} Mid-level representations simplify training, improve generalization, and aid training speed. \textbf{Mid-level representations should be the preferred input to policies}, especially for harder tasks, where they are more viable than alternative methods of invariance learning.

\noindent\textbf{Lessons learned:} 
    First, we found that agents trained using mid-level vision could be successfully trained for harder tasks than possible when training from scratch or using domain randomization. Training was less sensitive to the choice of hyperparameters, and training speed improved. Overall, these results ere consistent with the hypothesis that mid-level representations can simplify the input space and make the learning problem easier. 
    
    Second, we found the agents trained using mid-level representations were significantly more robust to domain shifts than agents trained from scratch (and also those trained using domain randomization). We showed this in simulation on multiple manipulation tasks under multiple types of domain shift (new objects, textures). We also showed this in the sim-to-real setting, successfully deploying mid-level-based simulator trained policies in unseen real-world buildings. The robust generalization is consistent with our second hypothesis, that mid-level representations also align training and test distributions to improve test-time performance. While approaches for solving mid-level objectives are generally less sensitive than methods trained for robotics using RL, they are still susceptible to domain shift. Improvements to methods for approximating these individual objectives would probably carry good knock-on effects for agents trained using mid-level vision.

    Third, which features performed well depended on the choice of task, but not so much on the environment used for training. The advantage of picking a good feature (vs. training from scratch) grew as tasks became more difficult, underscoring the importance of picking a good feature. While we did not explore how to pick a generic set of features, this would be an important avenue and an~\cite{sax2019midlevel} has proposed an initial (computationally-derived example). Without the dependence on task, these features would be expected to work well in most environments.
    
\textbf{Future work:} That mid-level features work well in harder contexts suggests that we are ready to ``close the loop'' between features and downstream tasks. Features are currently defined irrespective of any downstream task. Choosing a representative set of benchmark tasks that ``cover'' downstream robotic tasks would make it possible to choose new computer vision objectives that better benefit downstream motor tasks. This same strategy of a defining of benchmark tasks has made it possible to design ever-better architectures (e.g. ResNets in computer vision and Transformers in language) that work well for most perception tasks in that domain.




\acknowledgments{This material is based upon work supported by ONR MURI (N00014-14-1-0671), an Amazon AWS Machine Learning Award, NSF (IIS-1763268), a BDD grant and Toyota Research Institute (TRI)\footnote{\tiny{TRI provided funds to assist the authors but this article solely reflects the opinions and conclusions of its authors and not TRI or any other Toyota entity.}}.}


\scriptsize{\bibliography{main.bib}}

\end{document}